\documentclass[10pt,twocolumn,letterpaper]{article}

\usepackage{cvpr}
\usepackage{times}
\usepackage{epsfig}
\usepackage{graphicx}
\usepackage{amsmath}
\usepackage{amssymb}

\usepackage[utf8]{inputenc} 
\usepackage[T1]{fontenc}    
\usepackage{url}            
\usepackage{booktabs}       
\usepackage{amsfonts}       
\usepackage{nicefrac}       
\usepackage{microtype}      
\usepackage{graphicx}
\usepackage{bm}
\usepackage{subcaption}
\usepackage{amsmath}
\usepackage{xcolor}
\usepackage{wrapfig}
\usepackage{array}
\usepackage{color}
\usepackage{floatrow}
\usepackage{algorithm}
\usepackage{algorithmic}
\newfloatcommand{figurebox}{figure}[\nocapbeside][\dimexpr(\textwidth-\columnsep)/2\relax]
\newcommand{\tabincell}[2]{\begin{tabular}{@{}#1@{}}#2\end{tabular}} 
\usepackage{multirow}

\usepackage[pagebackref=true,breaklinks=true,letterpaper=true,colorlinks,bookmarks=false]{hyperref}

\cvprfinalcopy 

\def\cvprPaperID{4683} 

\ifcvprfinal\fi
\begin{document}

\title{Kernel Quantization for Efficient Network Compression}

\author{Zhongzhi Yu$^{2}$, Yemin Shi$^{1,3}$, Tiejun Huang$^{1}$, Yizhou Yu$^{3,4}$\\
$^{1}$ Department of Computer Science, School of EE\&CS, Peking University, Beijing, China \\ 
$^{2}$ Fu Foundation School of Engineering and Applied Science, Columbia University, New York, U.S.A. \\
$^{3}$ Deepwise AI Lab, Beijing, China \\
$^{4}$ Department of Computer Science, The University of Hong Kong, Hong Kong, China\\
{\tt\small zy2285@columbia.edu, \{shiyemin, tjhuang\}@pku.edu.cn, yizhouy@acm.org}
}

\maketitle
\thispagestyle{empty}

\begin{abstract}
	This paper presents a novel network compression framework, \textbf{Kernel Quantization} (\textbf{\textit{KQ}}), targeting to efficiently convert any pre-trained full-precision convolutional neural network (CNN) model into a low-precision version without significant performance loss. Unlike existing methods struggling with weight bit-length, \textit{KQ} has the potential in improving the compression ratio by considering the convolution kernel as the quantization unit. Inspired by the evolution from weight pruning to filter pruning, we propose to quantize in both kernel and weight level. Instead of representing each weight parameter with a low-bit index, we learn a kernel codebook and replace all kernels in the convolution layer with corresponding low-bit indexes. Thus, \textit{KQ} can represent the weight tensor in the convolution layer with low-bit indexes and a kernel codebook with limited size, which enables \textit{KQ} to achieve significant compression ratio. Then, we conduct a 6-bit parameter quantization on the kernel codebook to further reduce redundancy. Extensive experiments on the ImageNet classification task prove that \textit{KQ} needs 1.05 and 1.62 bits on average in VGG and ResNet18, respectively, to represent each parameter in the convolution layer and achieves the state-of-the-art compression ratio with little accuracy loss.
\end{abstract}
	
\section{Introduction}
In recent years, deep convolutional neural networks (CNNs) have achieved astonishing success in a variety range of computer vision tasks, such as image classification~\cite{krizhevsky2012imagenet,he2016deep}, semantic segmentation~\cite{AAAI_ref1}, action recognition~\cite{shi2017sequential}, and video restoration~\cite{AAAI_ref2,AAAI_ref3}. The promising results of CNNs are mainly attributed to the massive learnable parameters, which then benefit from abundant annotated data and computing platform improvement. Unfortunately, the increasing of learnable parameter consumes more memory and other computational resources, making it hard to deploy on resource-limited devices. According to \cite{denil2013predicting}, network parameters have significant redundancy, which inspired many works on network compression~\cite{han2015learning}. Among all of the network compression methods, network quantization attracts much attention for its ability in reducing the number of bits needed to represent each parameter. 

In network quantization, continuous parameters are mapped to a certain amount of discrete values (codebook). Thus, each parameter is represented by an index. However, network quantization still needs at least one bit to represent each parameter, leading to the theoretical compression ratio limit of 32 times. Impressive attempts have been made to achieve this limit in  ~\cite{rastegari2016xnor,courbariaux2015binaryconnect,hubara2016binarized}. In \cite{faraone2018syq}, the author proposed to use symmetric quantization and achieved the state-of-the-art performance among low-bit quantization methods, but still facing about 3\% accuracy loss on larger network structures such as ResNet18~\cite{he2016deep} and VGG~\cite{simonyan2014very}. Another approach is to train a low-bit network from scratch with some well-designed strategies. Zhang \textit{et al.}~\cite{zhang2018lq} proposed to train a neural network and learn quantizer with arbitrary-bit precision.

A severe problem along with reaching the theoretical compression ratio limit is the loss of accuracy. Almost all of these methods are suffering from significant accuracy loss, especially when using three or fewer bits on large-scale datasets (e.g., ImageNet~\cite{russakovsky2015imagenet}). Because of too few discrete values in the codebook, the convolution kernel lacks variety (the fewer entries in the codebook, the fewer combinations the kernels have).  To the best of our knowledge, for all of the conventional quantization methods, 1-bit quantization leads to significant accuracy loss and 2-bit quantization has relatively lower accuracy loss but with limited compression ratio of 16 times.

To overcome the dilemma of variety and the theoretical compression ratio, we first propose to consider the convolution kernel as the quantization unit to bind the theoretical compression ratio limit to the kernel size, which normally is $3\times 3$, instead of each parameter. Secondly, we propose to apply 6-bit quantization to the kernel codebook to further compress the model meanwhile preserve the variety of kernels. With these two steps, we are able to significantly compress the CNN without the limitation of theoretical compression ratio and achieve comparable accuracy to the full-precision model. For a $3\times 3$ kernel, we are able to increase the theoretical limit from 32 to 288.

In summary, our major contributions in this paper are as follows:

\vspace*{-.1cm}
\begin{itemize}
	
	\item The theoretical compression ratio limit of conventional network quantization method is introduced and analyzed, and a new perspective of both kernel-level and weight-level quantization is proposed to inspire others to break through the limitation.
	
	\item A novel method, Kernel Quantization (\textit{KQ}), is proposed to consider each kernel as a unit for kernel-level quantization and parameter quantization is then used to compress the kernel codebook.
	
	\item The proposed method is applied on popular CNN architectures and achieves significant compression ratio (on average 1.05 and 1.62 bits for VGG and ResNet18 to represent each parameter in the convolution layer, respectively) while having better accuracy compared to conventional network quantization methods.
	
\end{itemize}

The remainder of the paper is organized as follows. Section 2 presents related works on network quantization. Section 3 introduces the proposed \textit{Kernel Quantization} in detail and analysis the theoretical compression limits for conventional quantization methods. Section 4 provides the implement details, hyper parameter analysis and the experiment results. Section 5 concludes the paper.

\section{Related Works}

In 2016, ~\cite{han2015deep} proposed a quantization method which generates a codebook of discretization values using k-means clustering and maps all the parameters to the closest entry in the codebook. But the proposed method needs at least 4-bit for convolution layer and 2-bit for fully connected layer. ~\cite{zhou2017incremental} proposed an incremental network quantization method that performs weight partition, group-wise quantization and re-training in an iterative manner. But their method suffers significant accuracy loss when using 2-bit quantization. ~\cite{choi2016towards} derived that network quantization problem is related to entropy-constrained scalar quantization in information theory and designed a network quantization scheme that minimizes the performance loss with respect to quantization given a compression ratio constraint.

A different branch of network quantization is to train a low-bit network from scratch with some well-designed strategies~\cite{rastegari2016xnor}. ~\cite{hubara2016binarized} proposed to use binary weights and activations directly for computing the parameter gradients. This method is able to reduce memory size drastically, but results in significant accuracy loss on large datasets such as ImageNet. ~\cite{courbariaux2015binaryconnect} proposed to train BinaryConnect network with binary weights while keeping gradients as real values. Recently, ~\cite{faraone2018syq} have pushed the extremely low-bit network quantization record forward by a large margin. The author proposed to generate the codebook in a symmetrical manner. ~\cite{leng2018extremely} proposed to cast the original problem into several sub-problems. These method performs well on small networks like AlexNet, but suffers from non-negligible accuracy drop on larger models, e.g., VGG and ResNet18.

There are also works trying to exploit the benefit of representing multiple parameters with one index. \cite{Gong2014compressing} proposed to quantize fully connected layers in networks with vectors as quantization unit, but representing three parameters with one index did not provide promising compression ratio. In ~\cite{wu2018deepkmeans}, the author did quantization on each row of the convolution kernel. However, such method performs well on small datasets but suffers significant performance reduction in large datasets such as ImageNet~\cite{russakovsky2015imagenet}. In \cite{son2018clustering} the author explored further but they introduced enormous hyper parameters that closely connected to the quantization performance and failed in finding a framework to efficiently quantize CNN.

Among all these methods, extremely low-bit quantization is still an open problem and far from being solved, especially when using large networks and large datasets. In this paper, we focus on boosting the quantization performance and breaking through the theoretical compression ratio limit, then propose the Kernel Quantization.

\section{Methods}
\begin{figure*}[t]
	\centering
	\includegraphics[width=130mm]{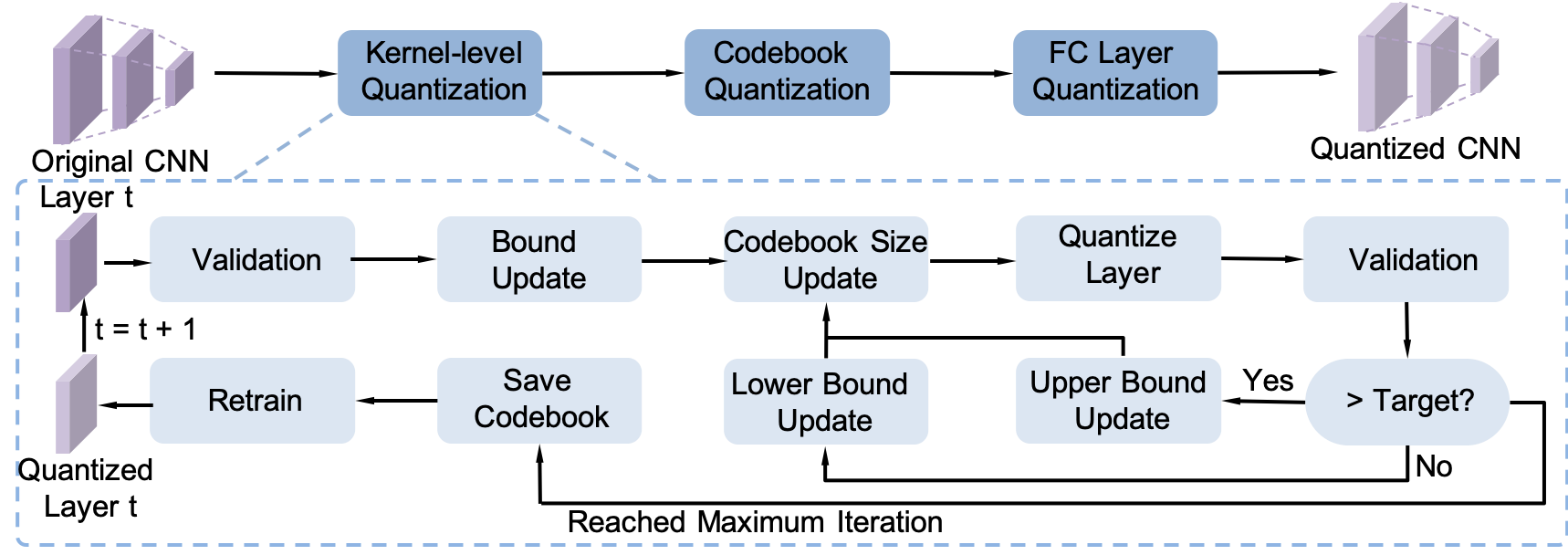}
	\caption{Illustration of Kernel Quantization framework. \textit{KQ} compresses the CNN with three steps, kernel-level quantization, codebook quantization and fully connected layer quantiztion. In the kernel-level quantization, we adopt a binary search based adaptive compression method, as shown in the light blue part in the figure. Given the initial entry ratio $\alpha$ and threshold ratio $r$, \textit{KQ} adaptively searches for the optimal codebook size for each layer, reducing the heavy burden of balancing the compression ratio and accuracy.}
	\label{fig:general} 
\end{figure*}
In this section, we provide the insight and detailed description of Kernel Quantization (\textit{KQ}) method. The overall framework is shown in Figure~\ref{fig:general}.

\subsection{Kernel-level Quantization}
\begin{figure*}[!t]
	\centering
	\includegraphics[width=120mm]{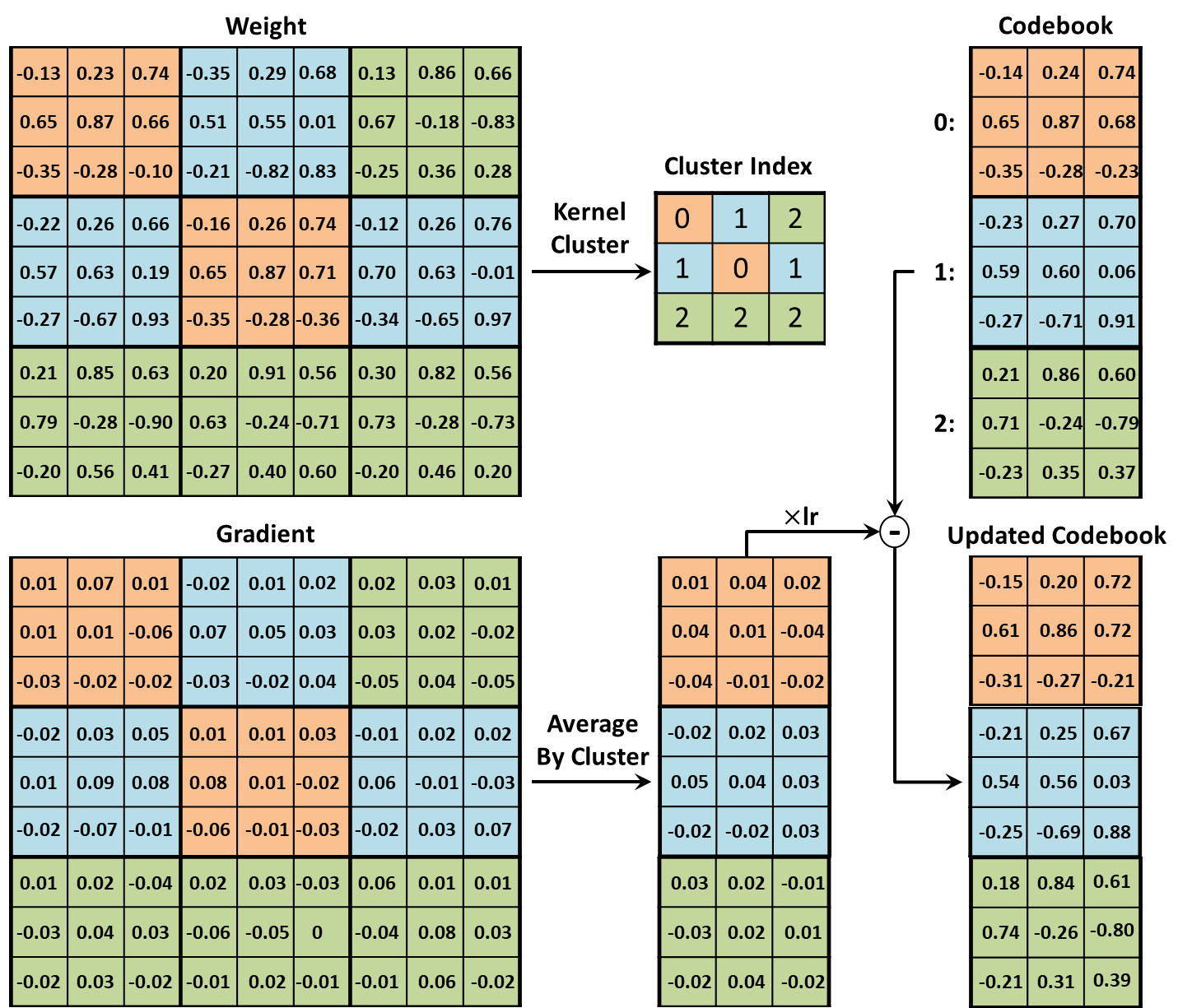}
	\caption{The pipeline of Kernel-Level Quantization and finetuning.}
	\label{fig:process} 
\end{figure*}

The pipeline of kernel-level quantization is shown in Figure~\ref{fig:process}. 
For a convolution layer with weight tensor $\mathcal{W}\in\mathbb{R}^{\omega\times \omega\times p\times q}$, where $\omega$ denotes the kernel size, $p$ and $q$ are input and output channel, respectively. 
We reshape it as a matrix $\bm{W} = [\bm{m}_0, \bm{m}_1, \cdots, \bm{m}_{n-1}]\in \mathbb{R}^{\omega^2\times n}$, where $n=pq$ denotes the total number of kernels in $\mathcal{W}$, $\bm{m}_i\in\mathbb{R}^{\omega^2}$ denotes the $i$-th kernel in $\mathcal{W}$. 
We generate a $k$ entry kernel codebook $\bm{C}=\{\bm{c}_0, \bm{c}_1, \cdots, \bm{c}_{k-1}\}$. 
The kernel set corresponding to entry $\bm{c}_i$ is $\bm{S}_i=\{\bm{m}_0^i, \bm{m}_1^i,\cdots, \bm{m}_{z-1}^i\}$ where $z$ is the number of kernels assigned to $\bm{c}_i$. 
We adopt k-means to minimize the following distance:
\begin{equation}
	\widehat{\bm{C}} = \arg\min_{\bm{C}} \sum_{i=0}^{k-1}\sum_{\bm{m}^i\in \bm{S}_i}\|\bm{m}^i-\bm{c}_i\|^2.
	\label{eq:kmeans}
\end{equation}
All kernels in $\mathcal{W}$ are mapped to corresponding entries in the kernel codebook. We represent each kernel with an index to the corresponding entry. Kernel-level quantization is performed in a layer by layer manner. During retraining, back propagation on codebook is done in a scheme like conventional quantization. We first calculate the gradient of each parameter in $\mathcal{W}$. Then, we calculate the elementwise average of gradients that are mapped to the same entry. The average gradients are used to update the kernel codebook values as follows,

\begin{equation}
	\bm{c}^{t+1}_i = \bm{c}^t_i - \gamma \times \frac{\sum_{\bm{m}^i\in \bm{S}_i}\frac{\partial L}{\partial \bm{m}^i}|_{\bm{m}^i=\bm{c}_i}}{z},
	\label{eq:bp_update}
\end{equation}
where $L$ is the network loss, $\bm{c}^t_i$ and $\bm{c}^{t+1}_i$ are values of entry $\bm{c}_i$ in the codebook after updating for $t$ and $t+1$ times, $\gamma$ is the learning rate. 

In conventional quantization methods, only the bit length $b$ needs to be carefully selected. Given $b$, the codebook size is usually set as $2^b$, because the size of the codebook is relatively small. But for kernel-level quantization, the codebook size is significantly larger. Carefully selecting the codebook size to balance accuracy and storage saves plenty of space. Thus, it is important to find the appropriate codebook size with adequate time complexity.

A naive approach is setting the codebook size to different orders of 2 in descending order until it reaches a certain threshold accuracy. This method makes full use of index bits while failed to select the appropriate size of codebook precisely. Fitting the codebook size to orders of 2 may either waste space to store redundant entries or lower the accuracy because of insufficient codebook size. Therefore, we propose a binary search approach searching for appropriate codebook size. We notice that network accuracy increases along with codebook size increasing. Thus, we set up a tolerable maximal accuracy drop for kernel-level quantization on each layer. Then we use a binary search to find the best codebook size that close to the target accuracy. In this way, simply given the parameters, the network adaptively finds the suitable codebook size and compression ratio. Thus, we precisely control the size of the codebook and the trade-off between the codebook size and network accuracy.

We define the initial entry ratio as $\alpha$ and the threshold ratio for target accuracy $r$. We test the reference accuracy $A_{ori}$ without quantization on the current layer. The initial codebook size is  $N_{init} = \alpha \times n$, and the corresponding baseline quantized accuracy is $A_{base}$. We define the target accuracy $ A_{target}$ as
\begin{equation}
A_{target} = A_{base} - (A_{ori} - A_{base}) \times r.
\label{eq:target_acc}
\end{equation}
Then we search for the codebook size with binary search to reach the target accuracy $A_{target}$. The upper bound of codebook size $B_{upper}$ is initialized as $N_{init}$ and lower bound $B_{lower}$ is initialized to zero. The testing codebook size is $N_{curr} = 0.5 \times (B_{upper} + B_{lower})$. Then kernel-level quantization generates a codebook with $N_{curr}$ entries and quantizes the layer to test the validation accuracy. If the validation accuracy is higher than $A_{target}$, we set $B_{upper} = N_{curr}$, else $B_{lower} = N_{curr}$. These steps are repeated until reaching the target accuracy $A_{target}$ or the maximum number of iteration. After finished binary search and quantized current convolution layer, the whole CNN is retrained for one epoch to finetune the model before quantizing next layer. 
The overall pseudo-code for the kernel quantization step is shown in Algorithm \ref{alg:kq}.

\begin{algorithm}[t]
\caption{Kernel-Level Quantization}
\label{alg:kq}
\begin{algorithmic}[1]
\REQUIRE~~\\
The T-layer full-precision CNN with weight matrix $\bm{W}_t\in\mathbb{R}^{\omega^2\times n_t}$ for each layer t;\\
Maximum number of iteration; \\
Threshold ratio $r$, entry ratio $\alpha$; \\
\ENSURE~~ \\
Quantized CNN including kernel codebook and indexes\\
\FOR{each layer $t \in \{1,2,\cdots,T\}$}
\STATE Test validation accuracy $A_{base}$;\\
\STATE $A_{curr} = A_{base}$;\\
\STATE Initial codebooks size is $N_{init}=N_{curr} = \alpha \times n_t$;\\
\STATE Upper bound and lower bound are $B_{upper} = N_{init}$ and $B_{lower} = 0$, respectively;\\
\STATE Calculate target accuracy $A_{target}$ with Eq.~\ref{eq:target_acc};\\
\WHILE{i < iteration and $A_{curr} \neq A_{base}$}
\STATE $i = i + 1$;\\
\STATE Recover the original weight $\bm{W}_t$;\\
\STATE Quantize kernel in $\bm{W}_t$ with k-means in Eq.~\ref{eq:kmeans};\\
\STATE Test validation accuracy $A_{curr}$;\\
\IF{$A_{curr}>A_{target}$}
\STATE Update upper bound $B_{upper} = N_{curr}$;\\
\ELSE
\STATE Update lower bound $B_{lower} = N_{curr}$;\\
\ENDIF
\ENDWHILE
\STATE Retrain the CNN for one epoch, for quantized layers, update parameter with Eq.~\ref{eq:bp_update};\\
\ENDFOR
\end{algorithmic}
\end{algorithm}

\subsection{Kernel Codebook Quantization}
We further compress the kernel codebook after quantized kernels on all convolution layers. The model storage after kernel-level quantization includes two parts, codebook and indexes. Total number of bits $B$ needed to store the quantized convolution layer is 
\begin{equation}
	B = k \times \omega^2 \times b_c + n\times \text{ceil}(\log_2k),
\end{equation}
where $b_c$ is the bit length for each parameter in the kernel codebook. When $k$ is large, storing entries in the codebook consumes massive space. So quantizing the kernel codebook provides us with additional compression ratio. We adopt a simple yet effective 6-bit parameter quantization method for codebook quantization.

We use the layer by layer strategy to preserve performance. First, we do k-means clustering on all parameters in the codebook. A small trick is that since different entries appear for different times in $\mathcal{W}$, the more times an entry appears, the more important it is. So we use the entry appearance time as the weight of the kernel and perform weighted k-means. This helps the algorithm to pay more attention to preserving important parameters. Retraining of codebook quantization follows the same scheme as kernel quantization. We update the parameters with the average gradients from different kernels mapped to the same entry. Retraining is conducted after quantizing codebook for every two layers and iterate for one epoch. After quantizing all codebooks of the network, we further run a 6-bit quantization on the fully-connected layer with the same layer by layer k-means clustering method.

It is worth to notice that \textit{KQ} does not add much extra burden on testing. In the testing phase, the procedure is the same as conventional quantization methods with extra mapping. \textit{KQ} first recovers the kernel codebook, then recovers the weight tensor $\mathcal{W}$ from kernel codebook. 

\subsection{Theoretical Compression Ratio Limit Analysis}
Conventional quantization methods use an index to map each parameter to an entry in the codebook. When the length of the index is shortened by reducing the size of the codebook, the total storage is reduced. The compression ratio of conventional quantization (denoted as $C_{con}$) is
\begin{equation}
	C_{con} = \frac{n\times \omega^2 \times b_{FP}}{u\times b_1 + n\times \omega^2 \times \text{ceil}(\log_2u)},
\end{equation}
where $b_{FP}$ is the bit length of a full precision parameter, which is 32-bit in most cases, $u$ is the size of the codebook, $b_1$ is the length of each entry in the codebook. We obtain the theoretical limit when there are only two entries in the codebook and each parameter is represented with 1-bit. The theoretical limit of conventional quantization is 32.

\textit{KQ} bypasses this limit by mapping each kernel, instead of each parameter, to an entry in the codebook. Thus, \textit{KQ} uses one index to represent nine parameters (for a $3\times 3$ kernel). The compression ratio of KQ (denoted as $C_{KQ}$) is
\begin{equation}
	C_{KQ} = \frac{n\times \omega^2 \times b_{FP}}{k\times \omega^2\times b_c + n \times \text{ceil}(\log_2k)}.
	\label{eq:compression}
\end{equation} 
When the limit is approached, $k$ is 2. The equation turns to be $C_{KQ} = \frac{\omega^2 \times b_{FP}}{n \times \text{ceil}(\log_2k)}$ ($k$ is too small compared to $n$), the theoretical compression ratio is $\frac{n\times 3\times3\times32}{n\times log_22}  = 288$. 

\section{Experiments}
To evaluate the performance of the proposed Kernel Quantization (\textit{KQ}) method, we perform experiments on the large scale benchmark image dataset: ImageNet (ILSVR2012)~\cite{russakovsky2015imagenet}. We apply \textit{KQ} on VGG~\cite{simonyan2014very}, ResNet18~\cite{he2016deep} and GoogLeNet~\cite{googlenet} architectures.

\subsection{Experimental setup}
We evaluate \textit{KQ} on the image classification task. All of the images are resized to $256\times 256$. The images are then randomly cropped to $224\times 224$ patches with mean subtraction and random flipping without any data augmentation. We report the top-1 accuracy on the validation set under single-center-crop testing.

We implement \textit{KQ} on PyTorch~\cite{paszke2017automatic} platform and the referenced full precision CNN models are from the torchvision package. During retraining, we use SGD optimizer with learning rate 0.001, momentum 0.9. In the experiments, we only conduct kernel quantization on kernels of size $3\times 3$. We adopt the Yinyang k-means~\cite{ding2015yinyang} to deal with massive samples and cluster centers. Yinyang k-means has exactly the same results compared with conventional k-means but provides a significant boost in clustering speed. 

For the sake of narrative, we always name the first $3\times 3$ convolution layer as conv1, the second $3\times 3$ convolution  layer as conv2 and so on. When computing the compression ratio, we count all convolution layers regardless of the kernel size. In the rest of this section, we use ''K'' for Kernel-Level Quantization, ``C'' for Codebook Quantization, and ''K+C'' for apply Codebook Quantization after Kernel-Level Quantization.

To better compare \textit{KQ} with other quantization methods, we use the average number of bits per parameter (denoted as $\beta$) as the measurement. It is defined as the total number of bits needed to represent the convolution weight divided by the total number of parameters in the convolution weight:
\begin{equation}
\beta = \frac{k\times \omega^2\times b_c + n \times \text{ceil}(\log_2k)}{n\times \omega^2}.
\end{equation} 
It is the reciprocal of compression ratio in Equation \ref{eq:compression} multiplied by 32 (full precision bits).
\begin{figure}[t]
	\centering
	\begin{tabular}{cc}
		\includegraphics[width=0.45\linewidth]{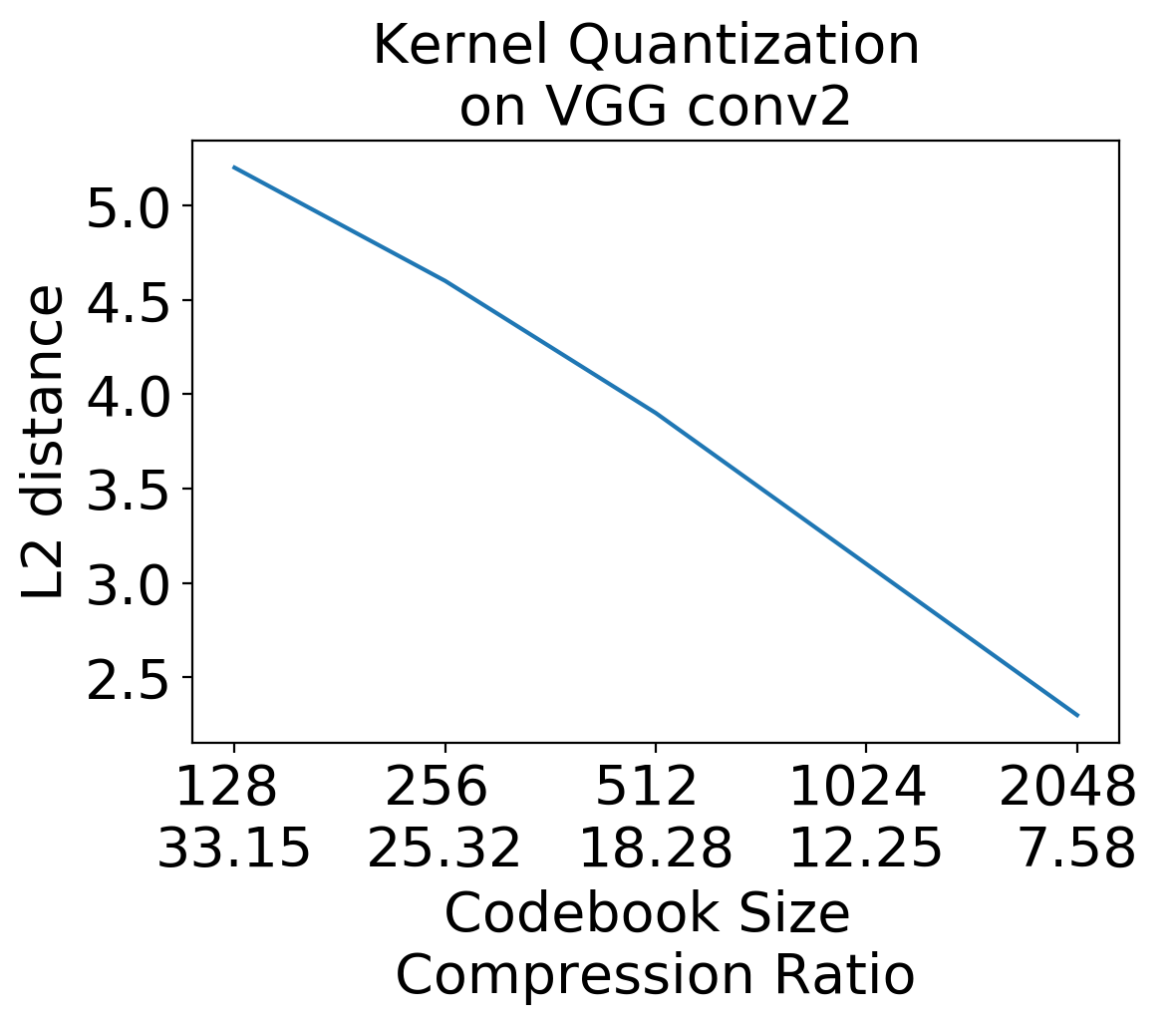}&
		\includegraphics[width=0.45\linewidth]{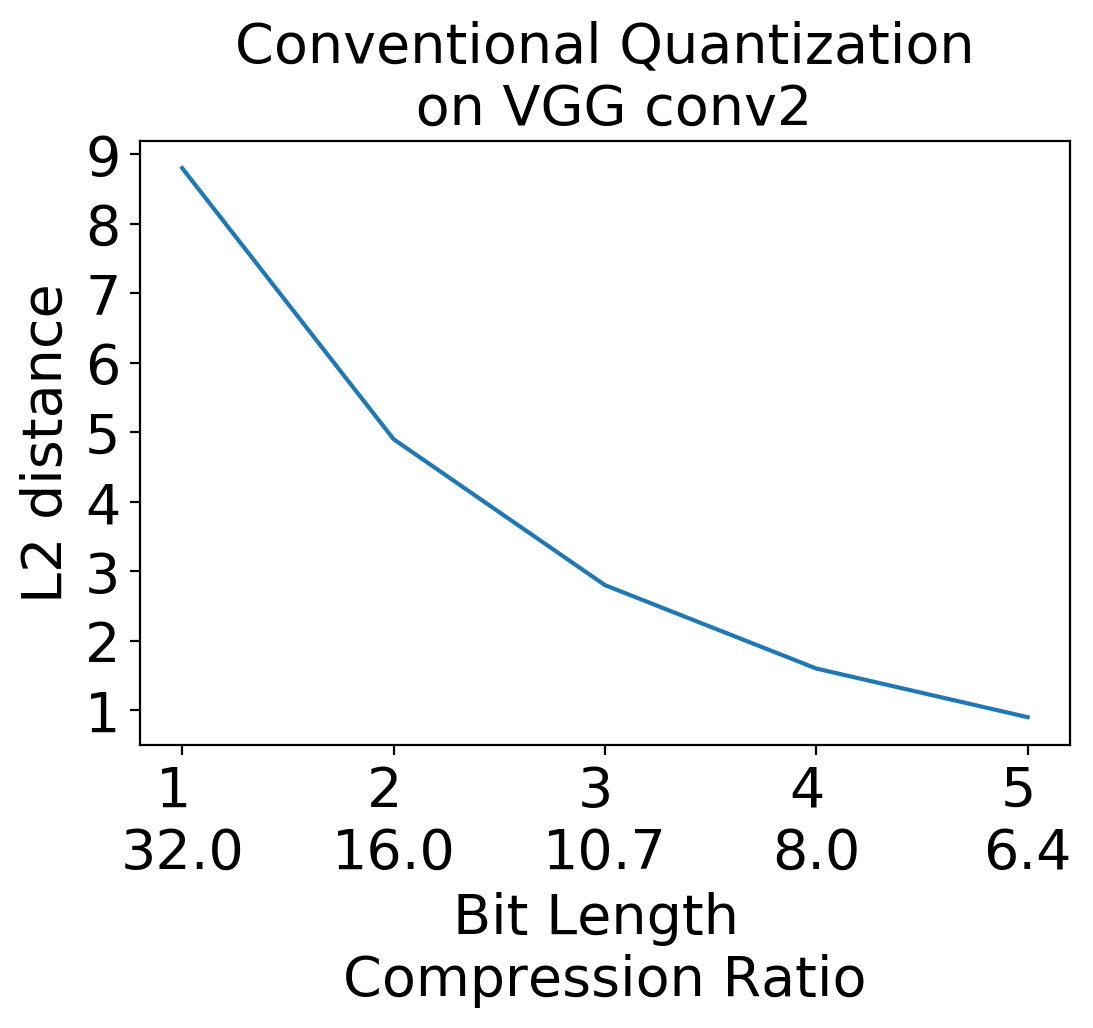}\\
		\includegraphics[width=0.45\linewidth]{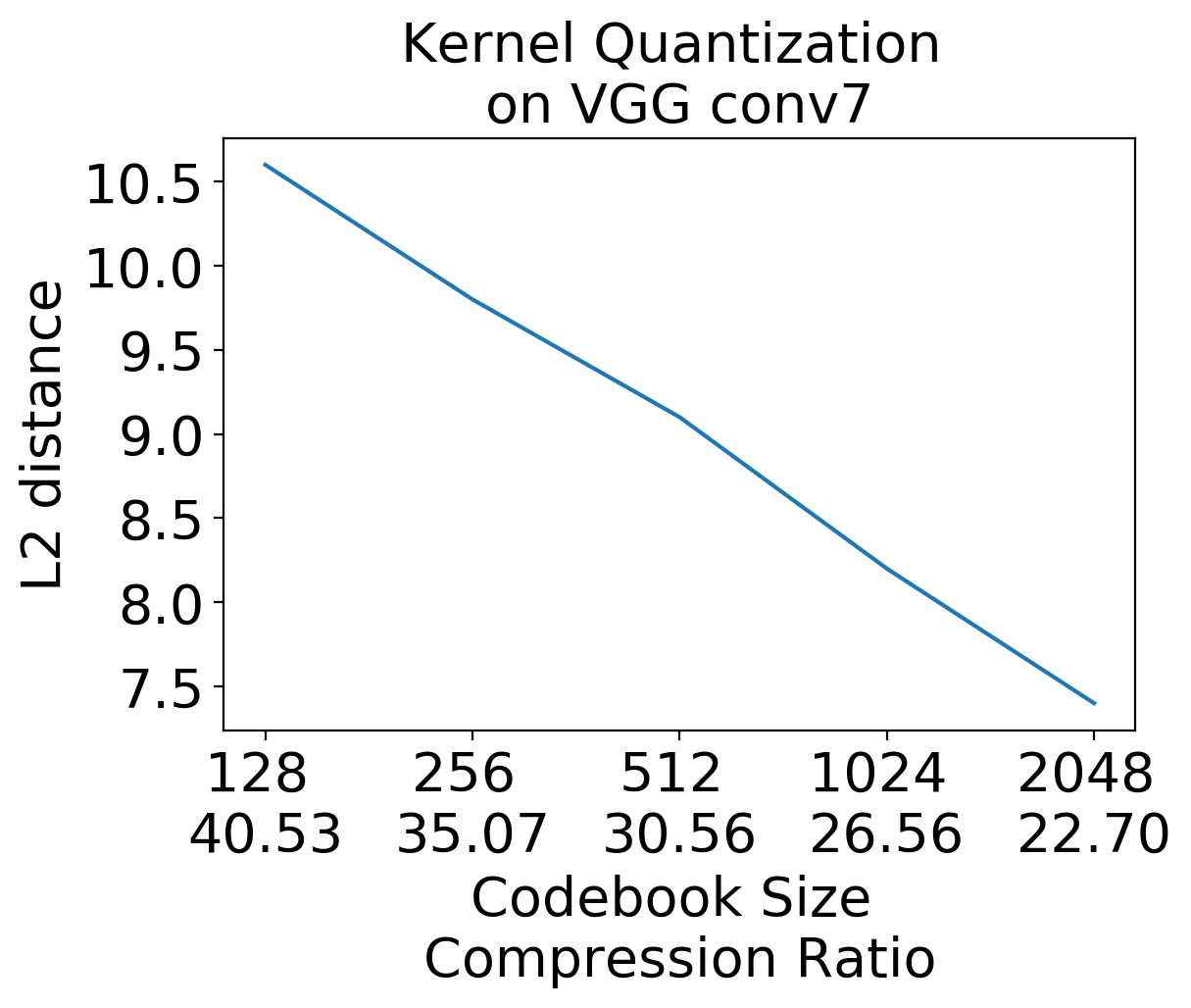} &
		\includegraphics[width=0.45\linewidth]{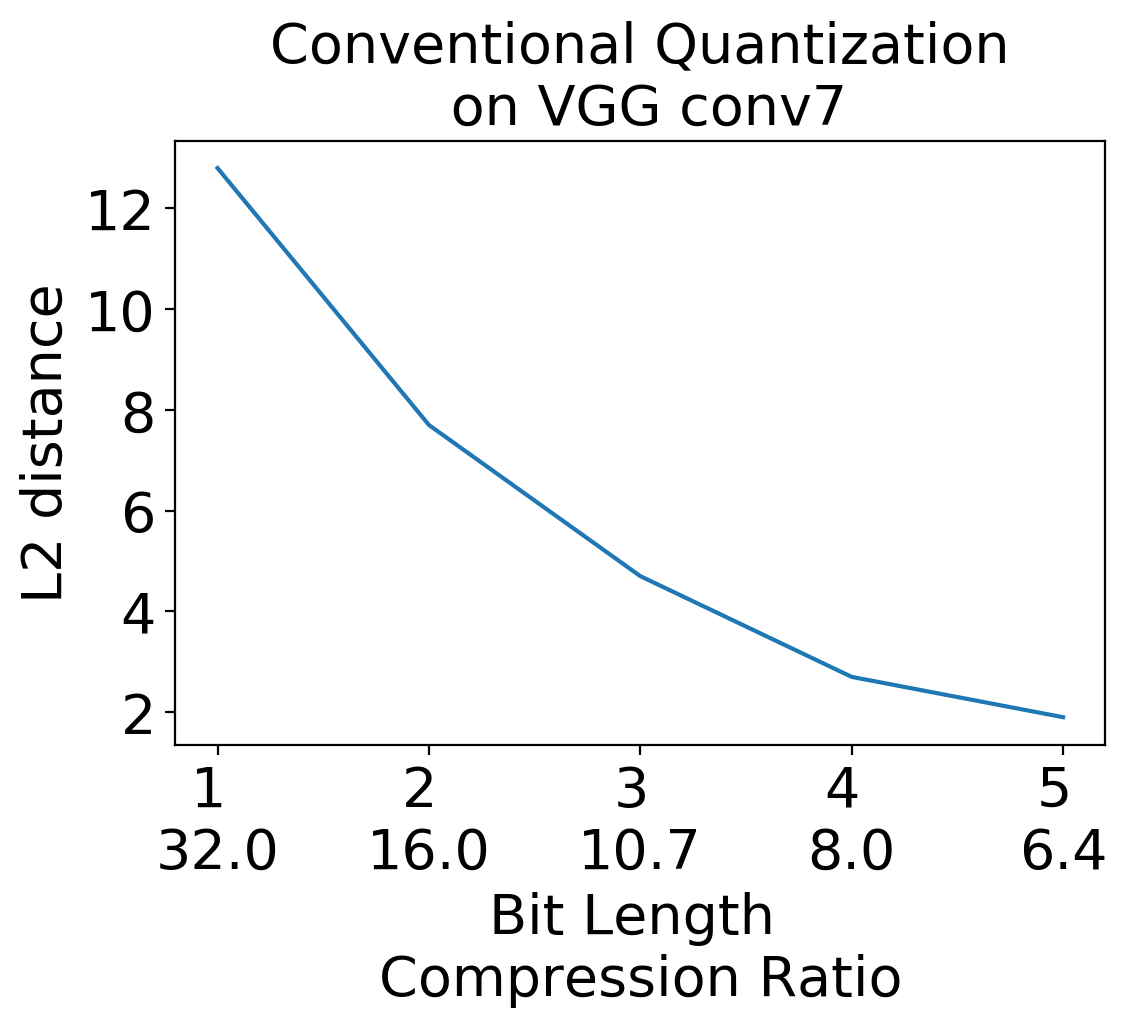}\\
		\includegraphics[width=0.45\linewidth]{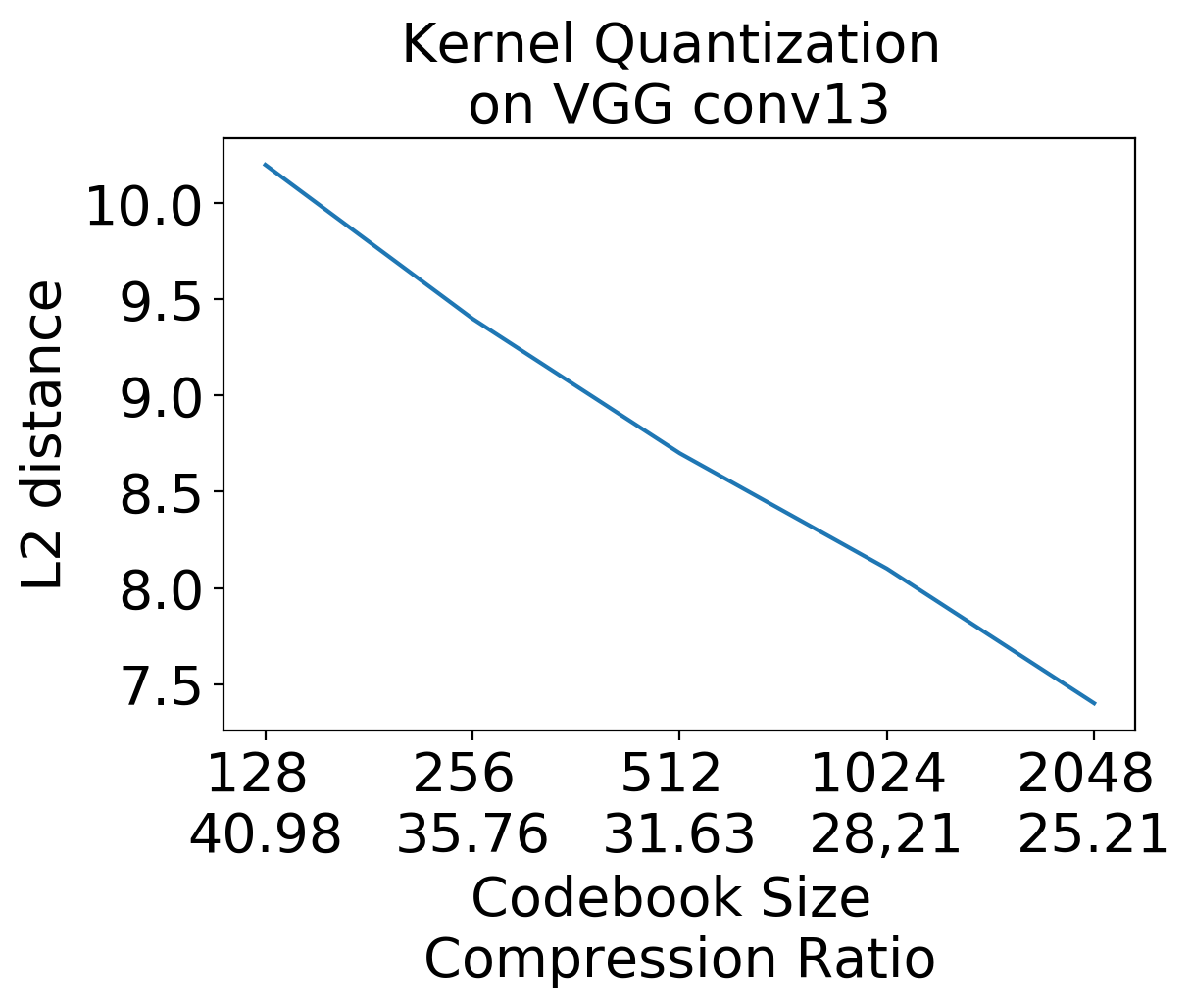}&
		\includegraphics[width=0.45\linewidth]{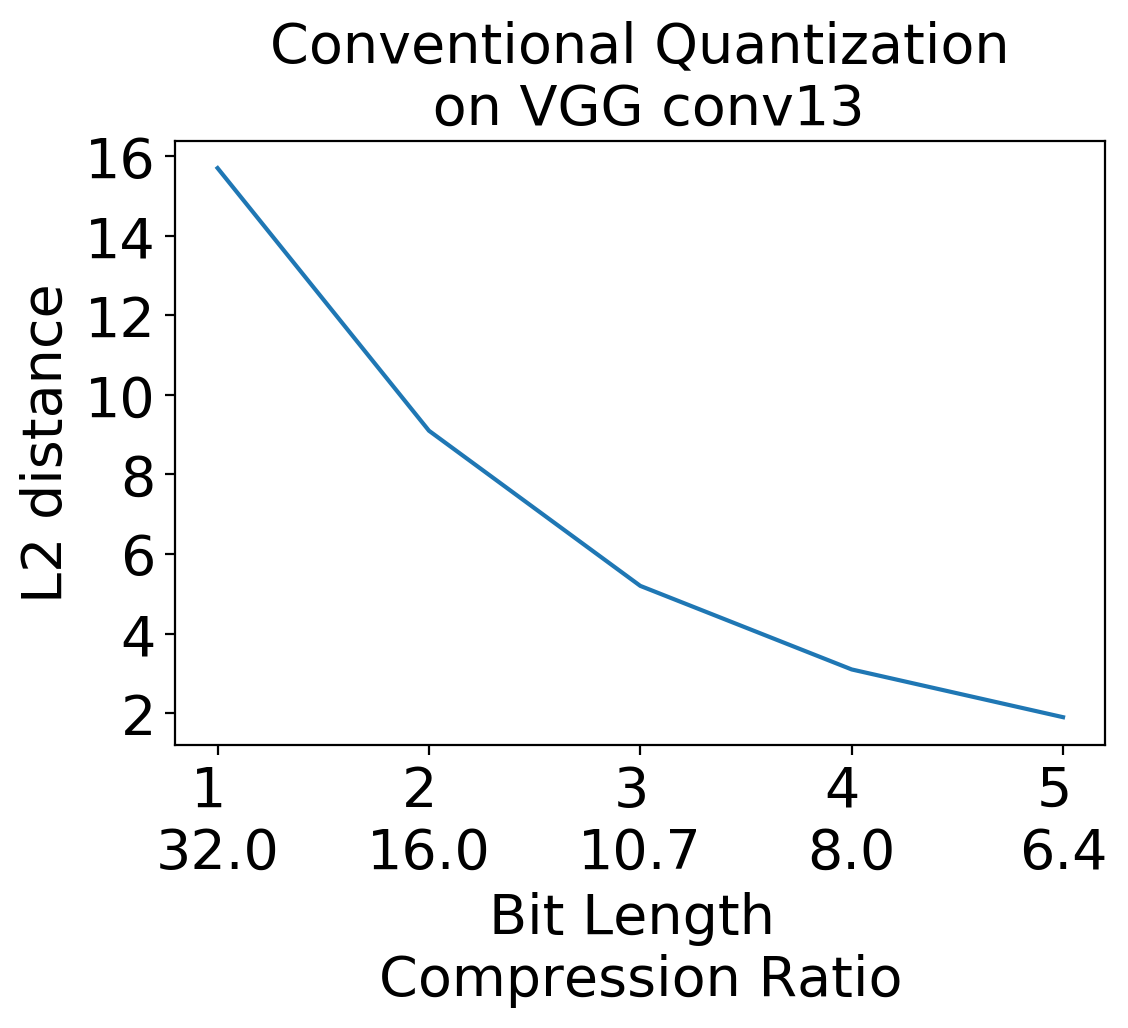}
	\end{tabular}
	\caption{$\ell_2$ distance between original weight and quantized weight under different codebook size or parameter bit length.}
	\label{fig:compare_others} 
	\vspace{-3mm}
\end{figure}
\subsection{Ablation Studies}
\subsubsection{Kernel Reconstruction Error Analysis}\label{subsec:loss_analysis}
Instead of finding the best match for each parameter in the codebook, kernel-level quantization only finds the kernel level best match. Theoretically, for a $\omega\times \omega$ kernel and a codebook for conventional quantization with $u$ entries, the codebook size needed for kernel-level quantization to represent all the possible combinations in the conventional quantization method is $k_{theoretical} = u^{\omega\times \omega}$. Network quantization aims to represent parameters with fewer bits. Thus, $u$ is a small integer in low-bit quantization. \cite{simonyan2014very} discovered that stacking smaller $3\times 3$ kernels obtains the same size of receptive field as a larger kernel. As the smaller kernel has fewer parameters and is computationally efficient, contemporary CNN architectures prefer smaller $3 \times 3$ kernels than larger kernels. Therefore, $k_{theoretical}$ is as small as $2^{3\times 3}=512$, making \textit{KQ} easily achieves equivalent, if not better than, representation ability to conventional extremely low-bit quantization methods. 

To exploit the reconstruction error of \textit{KQ} and conventional methods, we statistics the $\ell_2$ distance between the original weight and quantized weight in Figure \ref{fig:compare_others} under different codebook size or parameter bit length. The experiments are conducted on the 2nd, 7th, and 13th layers of VGG. We find among all three layers, \textit{KQ} with only 128 entries significantly outperforms conventional 1-bit quantization. Compression ratios of \textit{KQ} only are 18, 38.08, and 40.33, respectively. After applying codebook quantization, compression ratios are 33.15, 40.53, 40.98 and average numbers of bits per parameter are 0.965, 0.789, 0.780, respectively. \textit{KQ} with 256, 2048 and 512 entries on three layers outperform the conventional 2-bit quantization methods, respectively. Comparing with both 1-bit and 2-bit conventional quantization, \textit{KQ} achieves lower reconstruction error with higher compression ratio.

\subsubsection{Hyperparameter Analysis}
\begin{table}
    \centering
    \begin{tabular}{c|c|c|c|c}
    \toprule
         Threshold Ratio & 0.25 & 0.5 & 0.75 & 1.0 \\
         \hline
         Top-1 Accuracy & 70.1\% & 69.8\% & 69.4\% & 68.5\% \\
         \hline
         \tabincell{c}{Average Number\\of Bits} &  1.82 & 1.52 & 1.19 & 1.17\\
        \bottomrule
    \end{tabular}
    \caption{Relationship between threshold ratio $r$ and \textit{KQ} performance, initial ratio $\alpha$ is set to $0.5$. }
    \label{tab:threshold}
\end{table}
\begin{table}
    \centering
    \begin{tabular}{c|c|c|c|c}
    \toprule
         Initial Ratio & 0.25 & 0.4 & 0.5 & 0.6 \\
         \hline
         Top-1 Accuracy & 67.9\% & 69.2\% & 69.4\% & 69.5\% \\
         \hline
         \tabincell{c}{Average Number\\of Bits} &  1.12 & 1.19 & 1.19 & 1.23\\
        \bottomrule
    \end{tabular}
    \caption{Relationship between initial ratio $\alpha$ and \textit{KQ} performance, threshold ratio $r$ is set to $0.75$. }
    \label{tab:intial}
\end{table}
There are two key hyperparameters in \textit{KQ}, threshold ratio $r$ and initial ratio $\alpha$. We explore the effect of $r$ and $\alpha$ to the performance on VGG . The results are shown in Table \ref{tab:threshold}, \ref{tab:intial}. 

As the value of threshold ratio gets lower, the average number of bits turns higher. This is reasonable since given a fixed initial ratio $\alpha$, if the threshold ratio is lower, \textit{KQ} tends to compress the network more conservative with a higher target accuracy $A_{target}$. As shown in Fig \ref{fig:compare_others}, the relationship between $\ell_2$ distance and logarithm of codebook size is almost linear, so adjusting $r$ is an effectively way to precisely adjust the compression ratio and compressed network accuracy. 

The initial ratio $\alpha$ controls the least average bits per parameter \textit{KQ} achieves. By setting $\alpha=0.25$, after $8$ iterations in binary search, \textit{KQ} represents last several layers with a codebook with only $256$ entries. Thus the index for each kernel only consumes 8-bit. While for $\alpha=0.5$ and $\alpha=0.6$, \textit{KQ} generates a codebook with $512$ and $614$ entries in last several layers, using 9-bit and 10-bit to represent each entry in the codebook, respectively. The reason we prefer using $\alpha$ instead of iteration number is performing k-means clustering in deeper layers consumes more time. 

Since when $r=0.75$ and $\alpha=0.5$, \textit{KQ} provides the best trade-off between the accuracy and compression ratio, we adopt such settings in the experiment on VGG.

\subsection{Results on VGG}
VGG is widely used in a variety of computer vision tasks. It has 13 convolution layers and 3 fully connected layers connected in a sequential manner. All internal convolution layers (the first $7\times 7$ layer excluded) in VGG use $3\times 3$ convolution kernel. So we apply \textit{KQ} on all $3\times 3$ convolution layers. 
\begin{table}
	\caption{Comparison of \textit{KQ} with some of the state-of-the-art methods \cite{cai2017deep,zhang2018lq,wang2018two,faraone2018syq} on VGG. }
	\label{table:vgg_compare} 
	\centering
	\small
	\begin{tabular}{c|c|c|c|c}
		\toprule
		\hline
		Method &\tabincell{c}{Average\\ Number\\ of Bits}& Baseline & \tabincell{c}{Top-1\\ Accuracy} & \tabincell{c}{Accuracy\\ Loss}\\
		\hline
HWGQ & 2 & 69.8\% & 64.1\%  & 5.7\% \\
		LQ-Net  & 2 & 72.0\% & 68.8\%  & 3.2\% \\ 
		TSQ  & 2 & 71.1\% & 69.1\% & 2.0\% \\
		\hline
		LQ-Net & 1 & 72.0\% & 67.1\% & 4.9\% \\
		SYQ & 1 & 69.4\% & 66.2\% & 3.2 \% \\
		\hline
		ours(K) & \textbf{1.19} & \textbf{71.6\%} & \textbf{69.4\%} & \textbf{2.2\%} \\
		ours(K+C) & \textbf{1.05} & \textbf{71.6\%} &\textbf{69.2\%} & \textbf{2.4\%} \\	
		\hline
		\bottomrule
	\end{tabular}
\end{table}

As shown in Table \ref{table:vgg_compare}, we compare our method with some of the state-of-the-art methods. Kernel-level quantization itself is able to achieve on average 1.19 bits per parameter, which is about 26.9 times compression. Comparing with the methods which use approximately 2 bits per parameter, we outperform all the methods with fewer bits. As for LQ-Net~\cite{zhang2018lq} and SYQ~\cite{faraone2018syq} with 1 bit, \textit{KQ} uses similar bits while has much better performance than both methods. After further applying Codebook Quantization, our method is able to get 30.5 times compression with only another 0.2\% accuracy drop. Overall speaking, \textit{KQ} achieves a great balance between compression and accuracy, and outperforms all other methods.

\begin{table*}
	\centering
	\caption{Layerwise compression ratio of \textit{KQ} on VGG.}
	\label{table:vgg_layer_result} 
	\small
	\begin{tabular}{l|rr|rr|rr}
		\toprule
		\hline
		& \multicolumn{2}{c|}{Kernel} & \multicolumn{2}{c|}{Compression Ratio} & \multicolumn{2}{c}{Average Number of Bits} \\
		\hline
		Layer & Number & Codebook Size & K & K+C & K & K+C \\
		\hline
		conv1 & $3\times 64$ & 60 & 33.33\% & 7.94\% & 10.67 & 2.54\\
		conv3 & $64\times 128$ & 1008 & 15.80\% & 5.78\% & 5.05 & 1.84\\
		conv5 & $128\times 256$ & 1088 & 7.14\% & 4.44\% & 2.28 & 1.42\\
		conv7 & $256\times 256$ & 640 & 4.45\% & 3.65\% & 1.42 & 1.16 \\
		conv9 & $512\times 512$ & 512 & 3.32\% & 3.16\% & 1.06 & 1.01 \\
		conv11 & $512\times 512$ & 512 & 3.32\% & 3.16\% & 1.06 & 1.01 \\
		conv13 & $512\times 512$ & 512 & 3.32\% & 3.16\% & 1.06 & 1.01 \\
		\hline
		\bottomrule
	\end{tabular}
	\vspace{-3mm}
\end{table*}

To better understand how \textit{KQ} works on different layers, we list the layerwise compression ratio in Table \ref{table:vgg_layer_result}. An obvious observation is deeper layers has smaller kernel codebook than shallower layers. This observation, on the one hand, proves there is more redundancy in the deeper layers. On the other hand, it shows the shallower layers of VGG are more important, and network compression tasks exploit major compression ratio gain in deeper layers. Since the max iteration is set to 8, $\alpha=0.5$, and the last five layers all have $512\times 512$ kernels, the kernel codebook size of 512 presents that it is halved from $512\times 256$ consecutively for eight times and the validation accuracy is still above the target accuracy. This implies with more iterations in binary search,\textit{KQ} has potential to boost the performance further. 

\begin{figure}
	\centering
	\begin{tabular}{cccc}
		\includegraphics[width=0.45\linewidth]{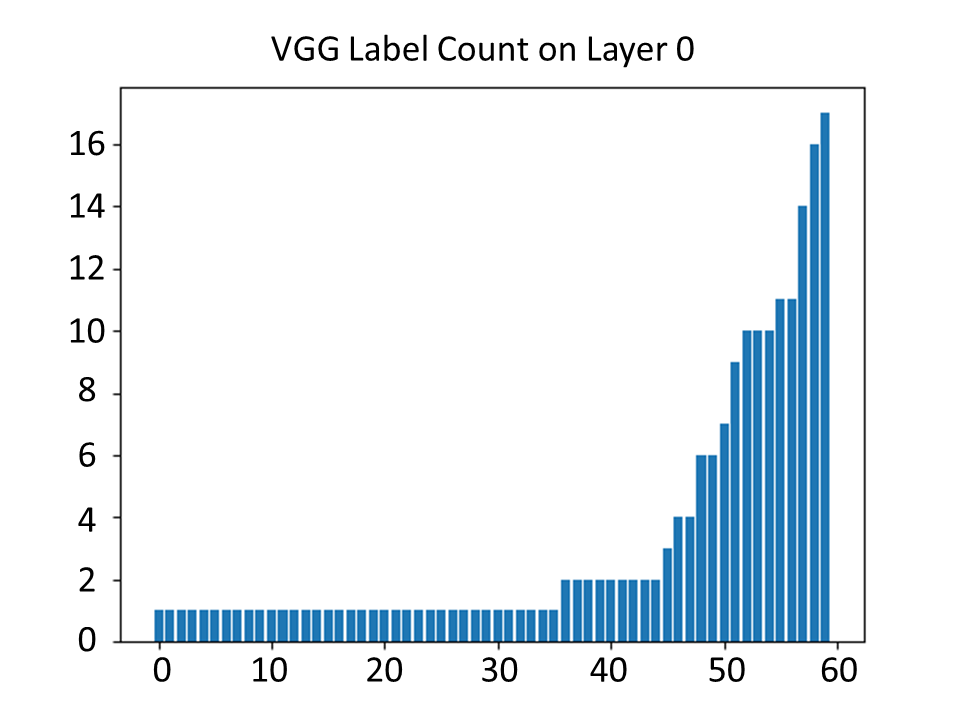}&
		\includegraphics[width=0.45\linewidth]{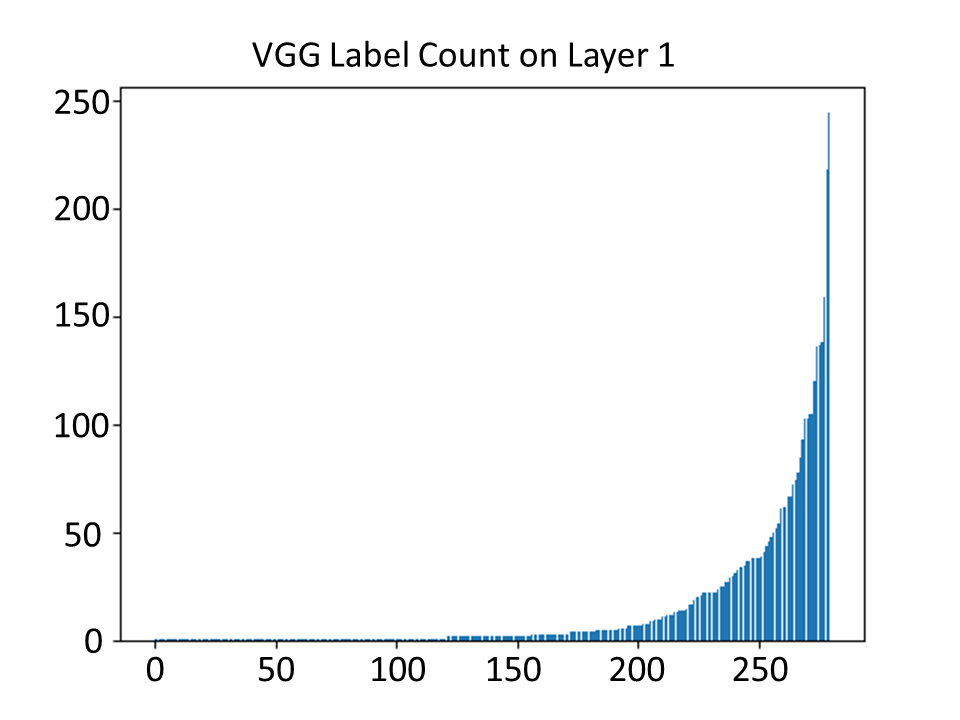}\\
		\includegraphics[width=0.45\linewidth]{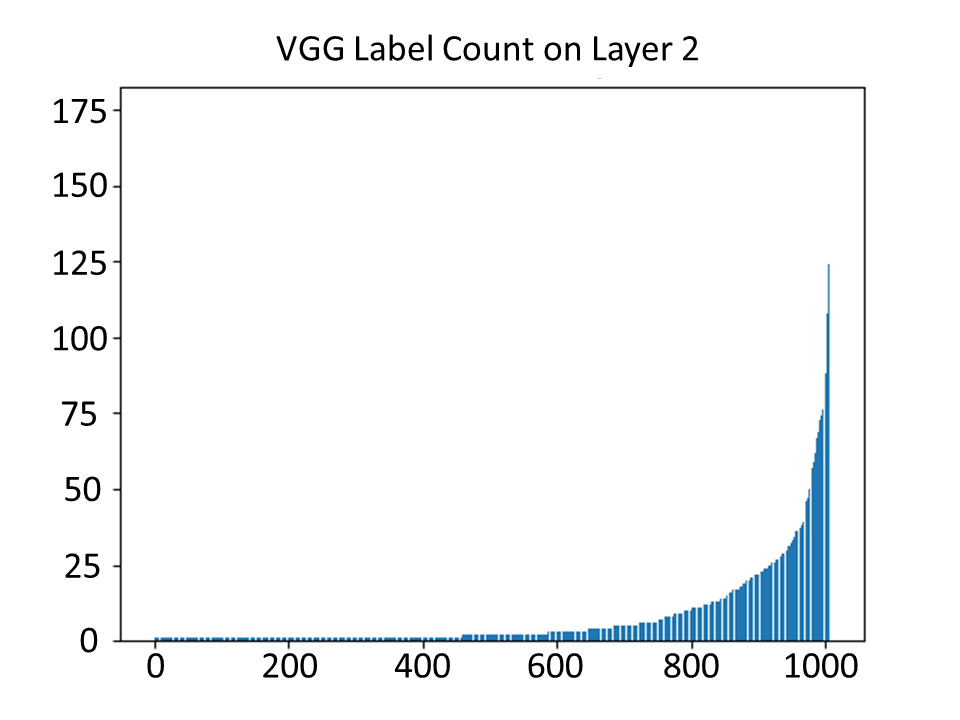}&
		\includegraphics[width=0.45\linewidth]{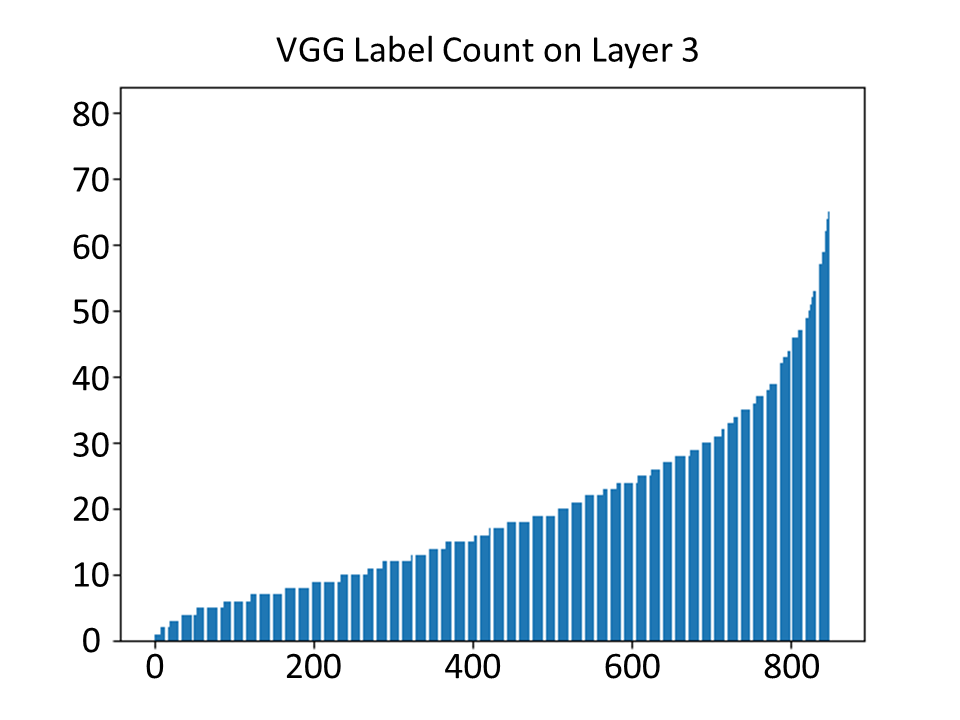}\\
	\end{tabular}
	\caption{Appearance statistics of each entry in the codebook on the first 4 layers of VGG.}
	\label{fig:vgg_label} 
	\vspace{-3mm}
\end{figure}

We further analysis the distribution of index after the kernel quantization. As shown in Figure \ref{fig:vgg_label}, we count the appearances of each entry in the codebook on the first 4 layers of VGG. It is clear that the distributions of the statistics are non-uniform, which makes it possible to further losslessly compress the network with coding methods like Huffman coding in \cite{han2015deep}.

\subsection{Results on ResNet18}
\begin{table*}
	\centering
	\caption{Layerwise compression ratio of \textit{KQ} on ResNet18.}
	\small
	\begin{tabular}{l|rr|rr|rr}
		\toprule
		\hline
		& \multicolumn{2}{c|}{Kernel} & \multicolumn{2}{c|}{Compression Ratio} & \multicolumn{2}{c}{Average Number of Bits} \\
		\hline
		Layer & Number & Codebook Size & K & K+C & K & K+C \\
		\hline
		conv1 & $64\times 64$ & 1203 & 33.28\% & 9.33\% & 10.64 & 2.98\\
		conv4 & $64 \times 64$ & 609 & 18.34\% & 6.26\% & 5.87 & 2.00\\
		conv7 & $128\times 128$ & 2974 & 22.31\% & 7.57\% & 7.15 & 2.42 \\
		conv10 & $256\times 256$ & 6987 & 15.17\% & 6.51\% & 5.01 & 2.15 \\
		conv13 & $256\times 512$ & 2303 & 5.92\% & 4.49\% & 1.95 & 1.48 \\
		conv15 & $512\times 512$ & 7679 & 7.44\% & 5.06\% & 2.38 & 1.62 \\
		\hline
		\bottomrule
	\end{tabular}
	\label{table:res_layer_result} 
	\vspace{-3mm}
\end{table*}
We also evaluate \textit{KQ} on ResNet18 architecture. Unlike the VGG, ResNet18 has batch normalization layers and skip connections directly connecting a deeper layer with a shallower layer to prevent the gradient vanishing. For ResNet18 network, we set $\alpha=0.3$, $r=0.5$, and the maximum iteration is 8.

\begin{table}[!htbp]
	\caption{Comparison of our Kernel Quantization method with some of the state-of-the-art methods \cite{zhou2017incremental,cai2017deep,zhang2018lq,faraone2018syq,rastegari2016xnor,yang2019quantization} on ResNet18. }
	\label{table:res_compare} 
	\centering
	\small
	\begin{tabular}{c|c|c|c|c}
		\toprule
		\hline
		Method & \tabincell{c}{Average\\ Number\\ of Bits} & Baseline &\tabincell{c}{Top-1\\Accuracy} & \tabincell{c}{Accuracy\\ Loss}\\ 
		\hline
		INQ & 3 & 68.3\% & 68.1\% & 0.2\% \\
		\hline
		INQ & 2 & 68.3\% & 66.0\% & 2.3\% \\
		HWGQ & 2 & 67.3\% & 59.6\%  & 7.7\% \\
		LQ-Net & 2 & 70.3\% & 64.9\%  & 5.4\% \\ 
		SYQ &2 & 69.1\% &68.0\% & 1.1\% \\
        QN & 2 & 70.3\% & 69.1\% & 1.2\% \\
		\hline
		XNOR-net & 1 &69.3\% & 51.2\% & 18.1\% \\ 
		LQ-Net & 1 & 70.3\%& 62.6\% & 7.7\% \\
		SYQ & 1 & 69.1\% & 62.9\% & 6.2\% \\
        QN & 1 & 70.3\% &66.5\% & 3.8\% \\
		\hline
		ours(K) & \textbf{2.99} & \textbf{69.7\%} &\textbf{69.0\%} & \textbf{0.7\%} \\
		ours(K+C) & \textbf{1.62} & \textbf{69.7\%} &\textbf{68.7\%} & \textbf{1.0\%} \\	
		\hline
		\bottomrule
	\end{tabular}
	\vspace{-3mm}
\end{table}

We compare \textit{KQ} with some of the state-of-the-art methods and show the results in Table \ref{table:res_compare}. Compared to INQ~\cite{zhou2017incremental}, although with only kernel-level quantization, INQ has less accuracy loss with the same average number of bits, but after applying the Codebook Quantization and further lower the average bit length, \textit{KQ} significantly outperforms the INQ with 0.38 fewer average number of bits and $1.3\%$ less accuracy loss. And \textit{KQ} has better performance with fewer bits comparing to other 2-bit quantization methods. Under 1-bit setting, \textit{KQ} achieves significant improvement with a little more bits.

In Table \ref{table:res_layer_result}, we report the layerwise compression ratio of \textit{KQ} on ResNet18. Comparing with VGG, we need larger codebook size on most of the layers. The most obvious difference is deeper layers of ResNet18 need larger codebook size than shallower layers. This is an aspect demonstrating the compactness of ResNet18. Meanwhile, \textit{KQ} achieves 1.62 bits per parameter without noticeable accuracy loss on such a compact network, proving the effectiveness of \textit{KQ}.

As the increasing of codebook size, we need more bits to represent each index. However, in \textit{KQ}, each index represents nine parameters. For 1-bit increase of index bit length, the average number of bits per parameter only increase by $\frac{1}{9}$ but the codebook size is doubled. With this merit, \textit{KQ} achieves an extremely low average number of bits per parameter on most large networks.

\subsection{Comparison with Structural Quantization}
\begin{table}[!htbp]
    \centering
	\begin{tabular}{c|c|c}
		\toprule
		\hline
		Method & Compression Ratio & Top-1 Acc Loss \\ \hline
		Deep k-Means & 2 & 0.17\% \\
		Deep k-Means & 3 & 0.32\% \\
		Deep k-Means & 4 & 1.95\% \\
		\hline
		GreBdec & 4.5 & 1.4\% \\
		\hline
		K+C & 5.78 & -0.2\% \\
		\hline
		\bottomrule
	\end{tabular}
	{
	\caption{Comparison of \textit{KQ}, GreBdec and Deep k-Means algorithm on GoogLeNet \cite{googlenet}}
	\label{table:deep_k_means} }
\end{table}
To further demonstrate the superiority of \textit{KQ}, we evaluate \textit{KQ} with the state-of-the-art structural quantization method deep k-means~\cite{wu2018deepkmeans} and structural compression method GreBdec~\cite{yu2017GreBdec} using GoogLeNet~\cite{googlenet} on ImageNet dataset. We set $\alpha=0.25$ and $r=1.0$ for \textit{KQ} on GoogLeNet. Table \ref{table:deep_k_means} shows the compression ratio of the above methods on convolution layers. Our method outperforms both of them with higher accuracy and compression ratio. This further proves the high efficiency of the proposed \textit{KQ} algorithm. Compared with deep k-means, we credit the superiority in performance to \textit{KQ}'s ability in preserving the tendency of the kernel, while~\cite{wu2018deepkmeans} proposed to use each row of the kernel as quantization unit failed to do so. As one of key functions of the convolution kernel is that it can extract the texture feature as a filter from the signal. The tendency of parameters in the convolution kernel is critical in preserving this ability. Given a row $[2,3,2]$ in kernel, it has the same distance to $[3,2,3]$ and $[1,2,1]$, but quantizing $[2,3,2]$ into $[1,2,1]$ is obviously better in preserving performance,  ~\cite{wu2018deepkmeans} failed in such case while our \textit{KQ} considers the kernel as a whole and better deals with these kinds of cases. Moreover, in~\cite{wu2018deepkmeans}, each index represents three parameters while in \textit{KQ}, each parameter represents nine parameters (for a kernel with size $3\times 3$). This provides a higher theoretical compression ratio for KQ.

\section{Conclusion}
In this work, we present a novel network quantization method, Kernel Quantization (\textit{KQ}), which aims to provide a highly efficient method with high compression ratio and low accuracy loss. By considering kernel as the quantization unit, \textit{KQ} boosts the theoretical limit of quantization from 32 to 288 and gives researchers more space for improvement. \textit{KQ} combines the kernel and weight level quantizations in a unified framework. The experiments prove that \textit{KQ} needs 1.05 and 1.62 bits for VGG and ResNet18 to represent each parameter and achieves the state-of-the-art compression ratio with little accuracy loss. Although \textit{KQ}'s improvement in performance is significant, there are still several future directions for research to exploit the potential of \textit{KQ} better. In our current implementation, we use k-means and relative accuracy change to determine codebook. However, this could be changed to methods like ~\cite{zhou2017incremental}, where kernels are quantized in descending order of importance to better preserve accuracy instead of minimizing kernel reconstruction loss. Moreover, we use k-means to further quantize codebooks and fully-connected layers. It could be possible to use lower bits with methods like ~\cite{faraone2018syq} to quantize codebook and fully-connected layers in a symmetric way. 


\bibliographystyle{ieee_fullname.bst}
\bibliography{ref} 

\end{document}